\documentclass{article}
\usepackage{spconf,amsmath,graphicx}
\usepackage{booktabs}

\usepackage{tikz}
\usepackage{textcase}
\usepackage{float}
\usepackage[pagebackref=true,breaklinks=true,letterpaper=true,colorlinks,bookmarks=false]{hyperref}

\title{DT-NeRF: Decomposed Triplane-Hash Neural Radiance Fields \\for High-Fidelity Talking Portrait Synthesis}

\name{Yaoyu Su, Shaohui Wang, Haoqian Wang\thanks{Corresponding author: wanghaoqian@tsinghua.edu.cn}}
\address{Shenzhen International Graduate School, Tsinghua University, Shenzhen 518071, China}

\begin{document}


%
\maketitle
%


\begin{abstract}

In this paper, we present the decomposed triplane-hash neural radiance fields (DT-NeRF), a framework that significantly improves the photorealistic rendering of talking faces and achieves state-of-the-art results on key evaluation datasets. Our architecture decomposes the facial region into two specialized triplanes: one specialized for representing the mouth, and the other for the broader facial features.  We introduce audio features as residual terms and integrate them as query vectors into our model through an audio-mouth-face transformer. Additionally, our method leverages the capabilities of Neural Radiance Fields (NeRF) to enrich the volumetric representation of the entire face through additive volumetric rendering techniques. Comprehensive experimental evaluations corroborate the effectiveness and superiority of our proposed approach.
\end{abstract}
\vspace{-0.7em} 
\section{Introduction}
\vspace{-0.3em} 
\label{sec:intro}
Audio-driven talking facial portrait synthesis represents a critical and challenging domain, particularly as augmented reality (AR), virtual reality (VR), and large language models (LLM) continue to find applications in 3D face-driven technologies such as digital humans, avatars, and remote conferencing. In recent years, researchers have extensively explored audio-driven facial synthesis in 3D visual settings \cite{chen2019hierarchical, prajwal2020wav2lip, thies2020nvp, zhou2020makelttalk, zhou2021pcavs, lu2021lsp, zhang2021facial} . Following the advent of Neural Radiance Fields (NeRF) \cite{mildenhall2021nerf} in 2020, this approach has been incorporated into this task, yielding impressive visual results. However, the original NeRF model has limitations in terms of computational speed and precise mouth synchronization during speech, indicating room for improvement.

NeRF \cite{mildenhall2021nerf}, a neural rendering technique, employs a 5D radiance field to capture complex 3D surfaces. Originally designed for rendering static, bounded scenes, the technology has since evolved to accommodate dynamic and unbounded settings.  NeRF has found applications in various sub-domains\cite{cao2023hexplane, fridovich2023k,  chen2022tensorf, fang2022fast}, such as scene reconstruction through the combination of Signed Distance Fields (SDFs) and volumetric rendering, facial and body rendering, and even hand reconstruction.

In the current arena of 3D facial synthesis leveraging Neural Radiance Fields  
 (NeRF)~\cite{wang2021multi, zhang2022learning},   prevalent methodologies tend to follow one of two paths. They either employ explicit 3D facial expression parameters or 2D landmarks, which can lead to a tangible loss of information especially the mouth area, or employ an implicit representation, using audio as a latent code to modulate or warp the canonical space. These strategies have deficiencies, especially in capturing the nuanced transformations of the mouth during speech.

\begin{figure}[t]
    \includegraphics[width=0.49\textwidth]{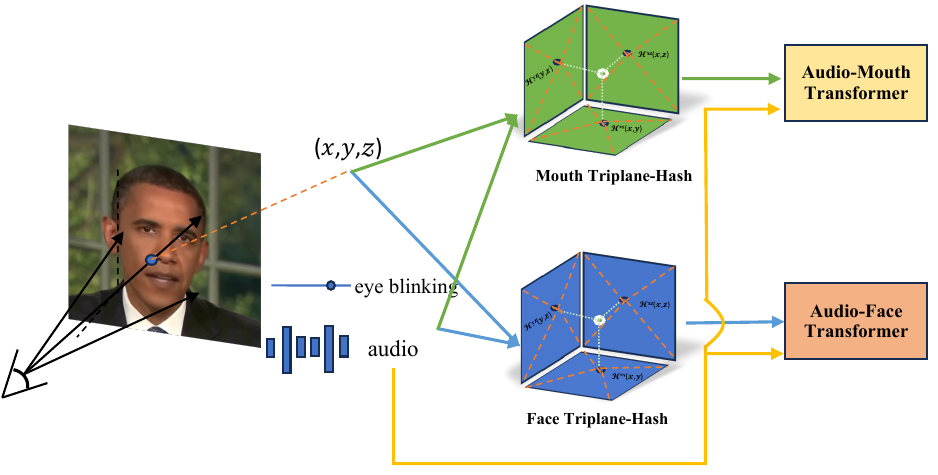}
    \vspace{-0.6cm}
    \caption{\small \textbf{Decomposed Triplane-Hash and Audio-Mouth-Face
     Transformer Module.}  Utilizing audio and eye-blinking features as conditional inputs, we modulate the decomposed triplane-hash of both the mouth and the face regions. Audio features are incorporated as residual terms and are used as query vectors in the transformer to align spatial points and their features.}
    \label{fig:my_label1}
    \vspace{-0.5cm}  
\end{figure}

Two main challenges are identified  in achieving high-quality, real-time audio-driven facial synthesis: 1) enhanced representation of the mouth and facial forms, and 2) effective coupling of audio signals with facial and mouth dynamics. To address these, we propose a two-fold approach. First, we employ a dynamic NeRF ~\cite{guo2021ad, liu2022semantic, shen2022dfrf, yao2022dfa}  that utilizes audio features as query of transformer aiming to optimize the density and color networks in NeRF to modulate from a canonical space to a dynamic space. Furthermore, we leverage the additive properties of color and volumetric density within the same NeRF space, thereby achieving a more seamless integration of audio and visual elements. 

\begin{figure*}[htbp]
  \centering
  \includegraphics[width=1\textwidth]{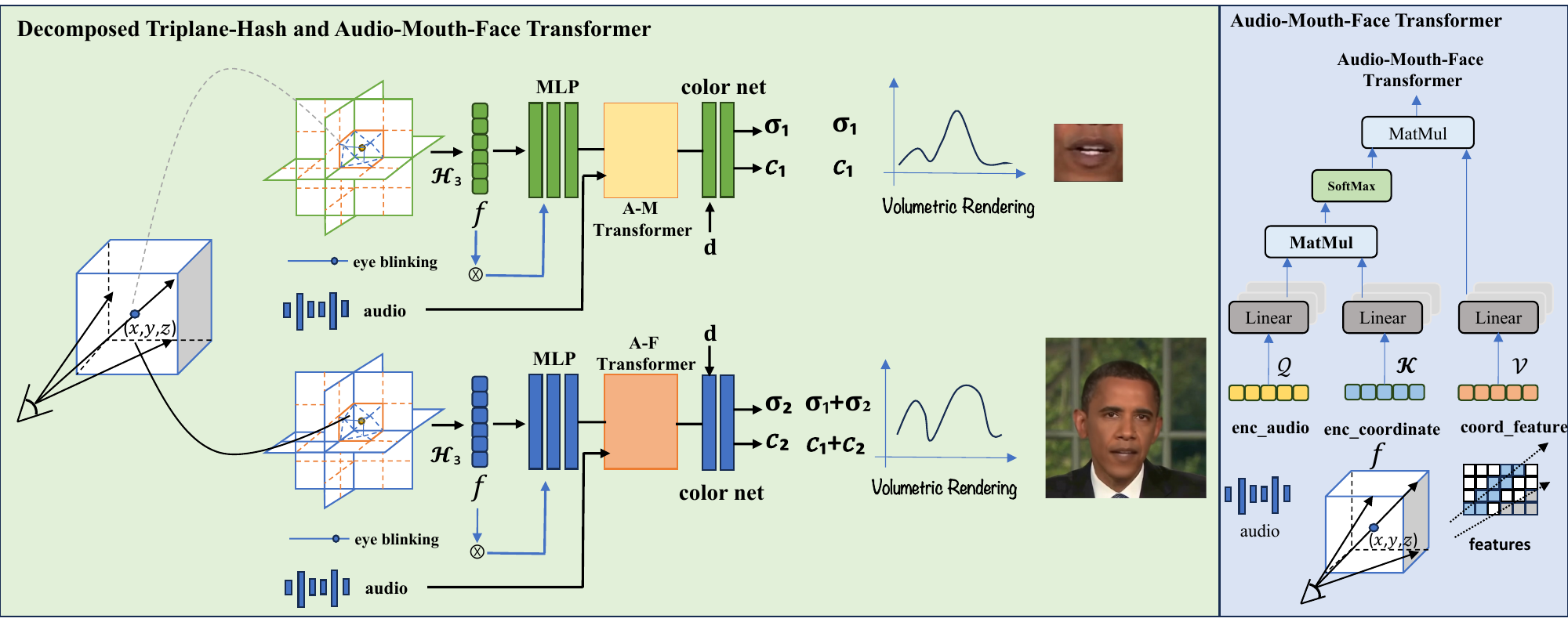}
  \vspace{-0.6cm}
  \caption{\small \textbf{Network Architecture.} Our model employ a decomposed triplane-hash representation for facial and oral regions, and utilize an Audio-Mouth-Face Transformer to treat audio features as queries for retrieving coordinates and corresponding attributes within a NeRF space. We then perform additive operations on the volumetric density ($\sigma_1$ , $\sigma_2$) and color attributes ($\mathbf{c}_1$ , $\mathbf{c}_2$) of mouth and facial representations sourced from the same NeRF space. After undergoing volumetric rendering, this results in an audio-driven speaking face.}
  \label{fig:my_label2}
  \vspace{-0.5cm}  
\end{figure*}

Our method aims to fuse audio cues more effectively with facial expressions, particularly focusing on the optimization of the mouth region. A decomposed 3D  representation is introduced by leveraging triplanes\cite{Chan2021} and audio-mouth-face transformer Module to align audio features with point-wise spatial characteristics, thereby enhancing the representational efficacy of audio-driven 3D facial synthesis. This allows for improved consistency and quality in the final, audio-driven 3D facial representation, as confirmed by evaluation metrics.

\vspace{-0.7em} 
\section{Methods and Descriptions}
\vspace{-0.3em} 
\label{sec:format}
\vspace{-0.7em} 
\subsection{Preliminaries and Problem Setting} \label{sec: preliminaries}
\vspace{-0.3em} 

NeRF \cite{mildenhall2021nerf} is employed to model the 5D plenoptic function, ranging from spatial coordinates $\mathcal{F}:(\mathbf{x}, \mathbf{d}) \rightarrow (\mathbf{c}, \sigma)$, where the 3D spatial coordinate  $\mathbf{x}=(x, y, z)$  and  viewing directions $\mathbf{d}=(\theta, \phi)$  , the density $\sigma$ and color $\mathbf{c}$.  NeRF operates by sampling along the rays $\mathbf{r}(t)$ to compute volumetric density $\sigma$ and color $\mathbf{c}$ attributes, using ray attenuation to measure the opacity in space. This differentiable rendering approach allows NeRF to reconstruct 3D scenes from 2D images and generate novel viewpoints.
\begin{equation}
    \setlength{\abovedisplayskip}{3pt}
    \setlength{\belowdisplayskip}{3pt}
    \hat{C}(r) = \int _{t_n}^{t_f}\sigma(\mathbf{r}(t)) 
    \cdot 
    \mathbf{c}(\mathbf{r}(t), \mathbf{d}) 
    \cdot
    T(t)dt , 
\end{equation}
where $t_n$ and $t_f$ are the near and far bounds. $T(t)$ is the accumulated transmittance from $t_n$ to $t$:
\begin{equation}
    \setlength{\abovedisplayskip}{3pt}
    \setlength{\belowdisplayskip}{3pt}
    T(t)=\mathrm{exp}(-\int_{t_n}^{t} \sigma(\mathbf{r}(s)) ds).
\end{equation}

Instant-NGP \cite{muller2022instant} employs hash encoding to represent sampled points in space and uses a multi-resolution framework to efficiently query volumetric density $\sigma$  and color $\mathbf{c}$ attributes, significantly accelerating the training process.
RAD-NeRF \cite{tang2022rad} introduce Instant-NPG \cite{muller2022instant}  as the equation,
\begin{equation}
    \mathcal{F^A}: (\mathbf{x}, \mathbf{d}, \mathbf{a}; \mathcal{H}) \rightarrow (\mathbf{c}, \sigma), 
\end{equation}
where $\mathcal{F^A}$ represents NeRF-based implict functions, $\mathbf{a}$ is audio features, $\mathcal{H}$ is multiresolution hash encoder.
ER-NeRF \cite{li2023ernerf} use the triplane-hash represation, which concat the hash-planes and get the represation:

\vspace{-1.3em} 
\begin{equation}
    \mathbf{f}_{\mathbf{x}} = \mathcal{H}^{\mathbf{XY}}(x,y) \oplus \mathcal{H}^{\mathbf{YZ}}(y,z) \oplus \mathcal{H}^{\mathbf{XZ}}(x,z).
\end{equation}
\vspace{-2.5em}

\begin{equation}
    \mathcal{F^B}: (\mathbf{x}, \mathbf{d}, \mathcal{D}; \mathcal{H}^3) \rightarrow (\mathbf{c}, \sigma),
\end{equation}

\noindent where  $\mathcal{D}$ is combined eye blinking features and audio features, $\mathcal{H}^3: \mathbf{x} \rightarrow \mathbf{f}_{\mathbf{x}}$ represents triplane-hash encoder.

\vspace{-0.7em} 
\subsection{Decomposed Triplane-Hash and Audio-Mouth-Face Transformer Module} \label{sec: preliminaries}
\vspace{-0.3em} 

Our proposed methodology decomposed the mouth and face representations through the triplane-hash representation \cite{li2023ernerf}, in Fig~\ref{fig:my_label1}. For improving the intricate details around the speaker's mouth, audio-driven modulation \cite{guo2021ad}  is applied specifically to the mouth area. This improves the overall performance of the Dynamic NeRF.


In our framework, visualized in Fig~\ref{fig:my_label2}, we employ OpenFace for the extraction of eye-blinking features as a means to enrich the facial feature set. These eye-blinking features are then integrated with the existing facial features and are processed through a Multi-Layer Perceptron (MLP) to produce spatial coordinates. These coordinates are further aligned with the output from a triplane-hash encoder. In conventional approaches, the first dimension of the density network output was employed for calculating density 
 $\sigma$, while subsequent dimensions were fed into a color network to obtain the color $\mathbf{c}$. However, such an approach has shown limitations in the ability of audio features to guide facial deformations effectively in canonical space to deformed space, particularly when the face undergoes significant deformations, causing challenges in spatial rendering and alignment.

Particularly, through visual analysis of deformations in the face during the audio-driven process, along with existing literature, it is evident that the impact of audio features on a talking face varies considerably \cite{li2023ernerf}. For instance, during speech, the region surrounding the mouth requires special attention for audio feature fitting, as it holds significant weight in performance metrics for the entire face. Based on these insights, we opt to decompose the face into the mouth region and other facial areas in Fig~\ref{fig:my_label2}. Both regions are then separately encoded using triplane-hash features in the same coordinate space.  

Our method introduces an audio-mouth-face transformer module in Fig~\ref{fig:my_label2}. In this architecture, audio features $\mathbf{a}$ act as query vectors in the transformer, while spatial coordinates $\mathbf{x}$  serve as key  vectors. The value vectors are composed of features $\mathbf{x\_feat}$  from spatial points fused with audio and eye-blinking features, and they are correspondingly aligned with features of the spatial points. The first dimension of the transformer's output serves as the density parameter 
 $\sigma$, while the remaining dimensions are processed through a color network to derive the color $\mathbf{c}$. Utilizing a mechanism akin to residual connections, along with an implicit mapping relationship between audio features and spatial points, we achieve a precise alignment of audio features with spatial coordinates. This significantly augments the capability of audio features to effectively drive deformations in the facial geometry.
\vspace{-0.3em} 
\begin{equation}
    \mathcal{F^C} : (\mathbf{a}, \mathbf{x}, \mathbf{x\_feat}; \mathcal{Q}, \mathcal{K},\mathcal{V}) \rightarrow (\mathbf{c}, \sigma),  
\end{equation}

\vspace{-0.3em} 

\vspace{-2.56em} 

\begin{align}
    \mathbf{q} &= \mathcal{Q}(\mathbf{a}), \\
    \mathbf{k} &= \mathcal{K}(\mathbf{x}),\\
    \mathbf{v} &= \mathcal{V}(\mathbf{x\_feat}), \\ 
    \mathbf{attn} &= \mathbf{Softmax}(q \cdot k) \cdot \mathbf{v},
\end{align}
\vspace{-1.7em} 

\noindent $\mathcal{F^C}$ denotes audio-mouth-face transformer mechanism. 

This approach enhances the robustness of our model, particularly in scenarios involving significant facial deformations, thereby improving the overall quality and reliability of audio-driven facial animation.



\begin{figure}[t]

    \includegraphics[width=1\linewidth]{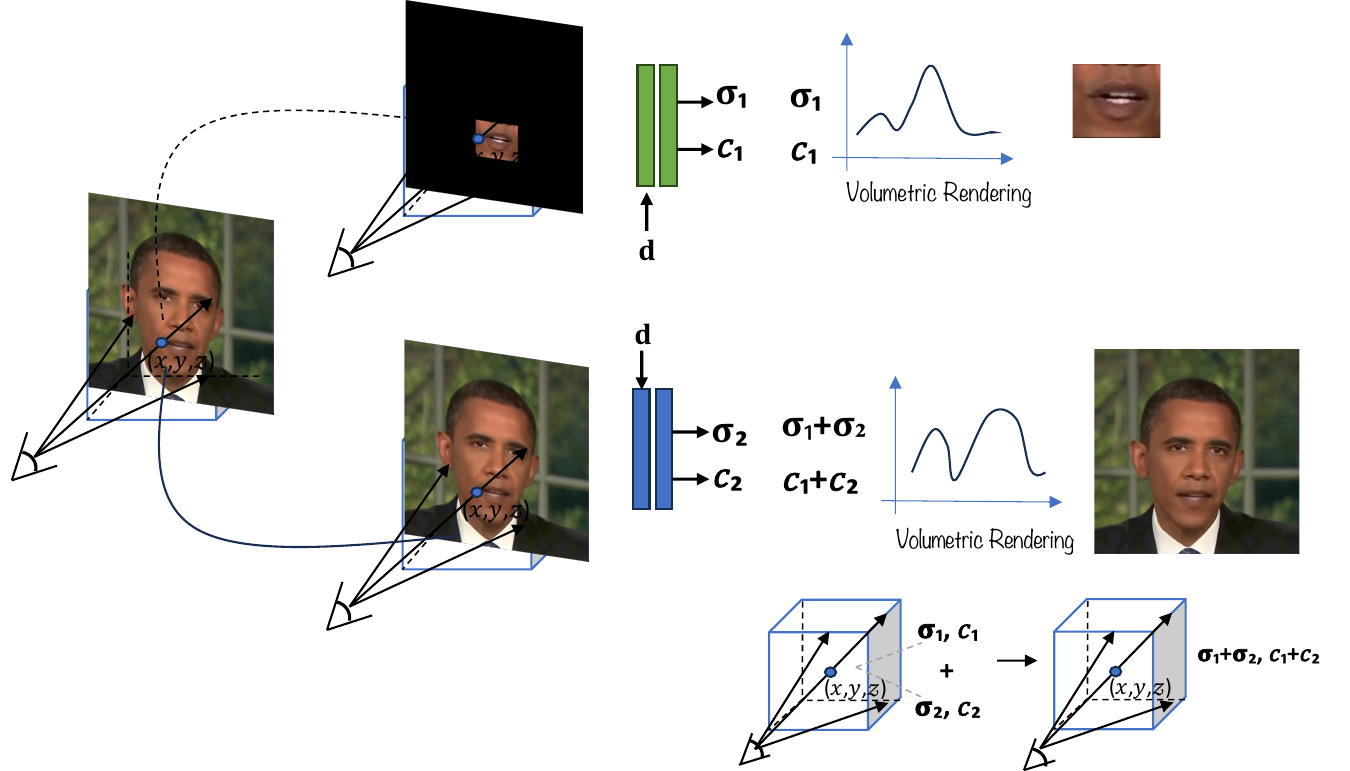}
    \caption{\small \textbf{Spatial Fusion in Volumetric Rendering.}  Facial densities $\sigma_{2}$ and colors $\mathbf{c_2}$ as well as the mouth densities $\sigma_{1}$ and colors $\mathbf{c_1}$ are additived sourced from the same NeRF space.}
    \label{fig:my_label3}
    \vspace{-1.5em}
    
\end{figure}

\vspace{-0.7em} 
\subsection{Spatial Fusion in Volumetric Rendering} \label{sec: preliminaries}
\vspace{-0.3em} 

Inspired by the NeRF-W \cite{martinbrualla2020nerfw}, which involves ray attenuation based on physical principles, we propose a spatial fusion of facial and mouth features after decomposing their representations in a unified coordinate frame in Fig~\ref{fig:my_label3}. During the training phase for the mouth region, we employ supervised learning using Mean Squared Error (MSE) Loss for pixel-level constraints. For full-face supervision, we integrate the separately enhanced mouth and face regions, thereby improving their collective representational capability.

Specifically, after the neural network training phase is complete, a spatial fusion technique is employed to seamlessly integrate the specialized training results for the mouth with those for the overall face. This technique merges the color and density maps derived from the triplanes corresponding to both the face and the mouth. Sampled from the same coordinate space and processed through decomposed triplane-hash encoding, the facial densities $\sigma_{2}$ and colors $\mathbf{c_2}$ along with the mouth densities  $\sigma_{1}$ and colors $\mathbf{c_1}$  are then summed to produce augmented facial information, which is subsequently subjected to two additional rounds of volumetric rendering.

This methodology enhances the learning process for deformations in the face and mouth from the audio features. Given that both the mouth and facial color and densities are sourced from the same coordinate space's triplane-hash encoding, their information correspondences align in coordinate space well. Through this mechanism, we have achieved results that are competitive such as PSNR, LPIPS, FID, thereby showcasing the effectiveness of this integrated approach for audio-driven facial modeling.

\vspace{-0.7em} 
\subsection{Training Details} \label{sec: preliminaries}
\vspace{-0.3em} 

Due to the incorporation of the decomposed triplane-hash and audio-mouth-face transformer module, we employ a pixel-level loss during the coarse stage of computation. For the facial region color , a MSE loss is calculated, followed by the computation of MSE loss for the mouth region color as well. In this context, $\hat{C}({\mathbf{r}})$ denotes the predicted color , and the value of $\lambda$ is set to 0.001.
\begin{small}
\begin{equation}
\setlength{\abovedisplayskip}{2pt}
\setlength{\belowdisplayskip}{2pt}
    \mathcal{L}_{coarse} = \sum_{\mathrm{i}\in\mathbf{}{I_{face}}}\left \| C({\mathrm{i}}) - \hat{C}({\mathrm{i}}) \right\|^2_2 + \lambda \sum_{\mathrm{j}\in\mathbf{}{I_{mouth}}}\left \| C({\mathrm{j}}) - \hat{C}({\mathrm{j}}) \right\|^2_2.
\end{equation}
\end{small}
\vspace{-0.6em}

After undergoing the coarse stage training duration as set in AD-NeRF \cite{guo2021ad}, we proceed to fine-tune specifically for the mouth region. At this stage, we employ a MSE loss and LPIPS loss targeting the mouth area . Following the settings from prior work, we set a set of patch  $\mathcal{P}$ ,  $\lambda$ is set to 0.001.
\begin{small}
\begin{equation}
\setlength{\belowdisplayskip}{3pt}
    \mathcal{L}_{fine} = \sum_{\mathrm{i}\in\mathcal{P}}\left \| C({\mathrm{i}}) - \hat{C}({\mathrm{i}}) \right\|^2_2 + \lambda\ \text{LPIPS}(\hat{\mathcal{P}}, \mathcal{P}).
\end{equation}
\end{small}

\vspace{-0.7em} 
\section{Experiments and Results Analysis }
\vspace{-0.3em} 
\label{sec:pagestyle}

\vspace{-0.7em} 
\subsection{Implementation details} \label{sec: preliminaries}
\vspace{-0.3em} 

We leverage the AD-NeRF \cite{guo2021ad} Obama dataset, comprising 3 to 5 minutes of 25fps video at a resolution of 450x450 pixels for the target subject. The training adopts a coarse-to-fine approach, initially focusing on the facial region, subsequently fine-tuning the mouth area, and finally extending to the attire. The baseline is  NeRF-based model such as AD-NeRF 
 \cite{guo2021ad}, RAD-NeRF \cite{tang2022rad}, and ER-NeRF \cite{li2023ernerf}.  The model is trained for approximately 2.5 hours on a single RTX 3090 GPU using the Adam optimizer in PyTorch. For quantitative evaluation, we employ metrics such as Peak Signal-to-Noise Ratio (\textbf{PSNR}), Learned Perceptual Image Patch Similarity (\textbf{LPIPS}), landmark distance  (\textbf{LMD}) , Fréchet Inception Distance (\textbf{FID}) .


    

\vspace{-0.7em} 
\subsection{Quantitative Evaluation} \label{sec: preliminaries}
\vspace{-0.3em} 
The results of Obama Dataset is as below,
\vspace{-0.3em} 

\begin{figure}[h]

    \includegraphics[width=1\linewidth]{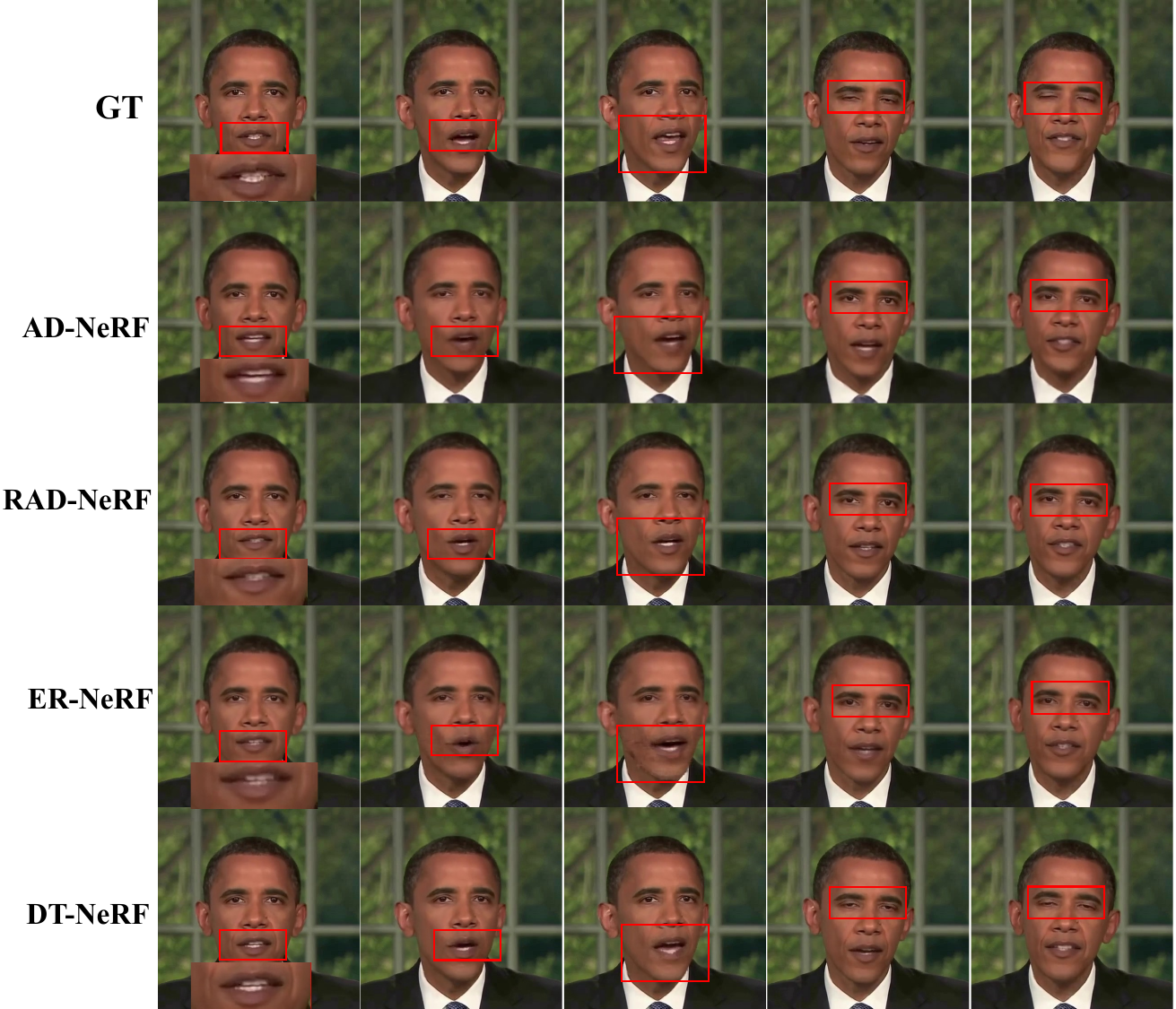}
    \vspace{-2.3em}
    
    \caption{\small \textbf{The comparison of the key frames and details of generated portrait.} Our approach in Obama Dataset \cite{guo2021ad} demonstrates significant advantages in terms of lip finesse, facial movements and eye blinking.}
    \label{fig:my_label4}
    
\end{figure}

\vspace{-1.9em} 
\begin{table}[h]
\begin{flushleft}  

\resizebox{1\linewidth}{!}{  
        \setlength{\tabcolsep}{2.5mm}  
        
        \begin{tabular}{lccccccccc}
        \toprule
        Methods & PSNR $\uparrow$ & LPIPS $\downarrow$ & FID $\downarrow$ & LMD $\downarrow$ & Time & FPS  \\
        Ground Truth  & N/A            & 0               & 0              & 0                      & -   & -       \\ \midrule
        AD-NeRF \cite{guo2021ad} & 30.75 &  0.1034 & 24.514 & 3.345 &18h &  0.08\\
        RAD-NeRF \cite{tang2022rad}     & 34.00    & 0.0387          &  10.835       & 2.696                   & 5h  & 32    \\ 
        
        ER-NeRF \cite{li2023ernerf}  & 35.37  & 0.0185 & 9.675 & 2.604       & \textbf{2h}  & \textbf{34}      \\ 
        DT-NeRF(Ours) & \textbf{35.39}   & \textbf{0.0169} & \textbf{9.472} & \textbf{2.601} & 2.5h & 32 \\
        \bottomrule 
        
        \end{tabular}
    }
    \setlength{\abovecaptionskip}{0cm}
    
    \vspace{-0.8em} 
    \caption{\small \textbf{The quantitative results of the \emph{head reconstruction setting}}.In obama dataset \cite{guo2021ad} under the resolution of $450\times450$.}
    \label{tab:setting1}

\end{flushleft}  
\end{table}

\vspace{-1.3em} 

To validate the generalizability of our approach, we conducted comparative experiments using a distinct speaking scenarios featuring individual sourced from YouTube.  Specially, in terms of posture, significant variations are observed.

\begin{figure}[h]

    \includegraphics[width=1\linewidth]{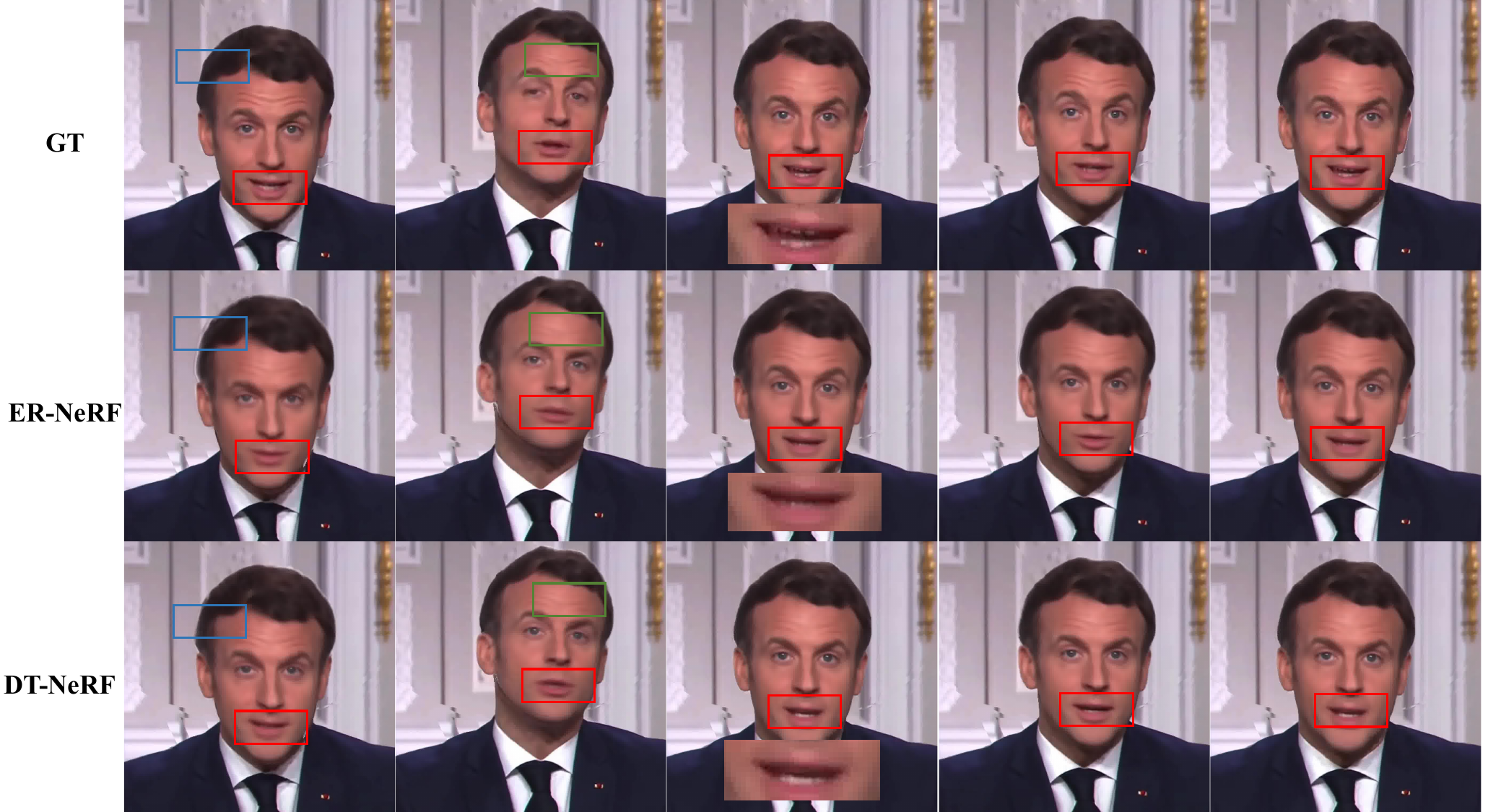}
    \vspace{-1.8em}
    
    \caption{\small \textbf{The comparison of the key frames and details of generated portrait.} Our approach in Youtube Dataset demonstrates significant advantages in terms of lip finesse, facial movements.}
    \label{fig:my_label5}
    \vspace{-0.2em} 
    
\end{figure}

\begin{table}[h]
\begin{flushleft}  

\resizebox{1\linewidth}{!}{  
        \setlength{\tabcolsep}{2.5mm}  
        
        \begin{tabular}{lccccccccc}
        \toprule
        Methods & PSNR $\uparrow$ & LPIPS $\downarrow$ & FID $\downarrow$ & LMD $\downarrow$ & Time & FPS  \\
        Ground Truth  & N/A            & 0               & 0              & 0                      & -   & -       \\ \midrule

        ER-NeRF \cite{li2023ernerf}     & \textbf{30.80}    & 0.054       & 12.110        & 5.54   & \textbf{2h}  & \textbf{34}    \\ 
        
        DT-NeRF(Our) & 30.45   & \textbf{0.048} & \textbf{11.274} & \textbf{5.34} & 2.5h & 32 \\
        \bottomrule 
        
        \end{tabular}
    }
    \setlength{\abovecaptionskip}{0cm}
    \vspace{-0.8em} 
    
    \caption{\small \textbf{The results from Youtube video.} Dataset Emmanuel Macron whose dataset under the resolution of $512\times512$.}
    \vspace{-1.8em} 
    \label{tab:setting1}
\end{flushleft}  
\end{table}

\subsection{Ablation Study} \label{sec: preliminaries}
\vspace{-0.3em} 

To verify the efficacy of our approach, we conducted ablation studies with the following experimental validations. When omitting the fusion of spatial points, the metrics were as follows. In experiments where spatial point fusion was employed but without the use of the transformer, the results yielded the following outcomes.

\vspace{-1em} 
\begin{table}[h]
\begin{flushleft}  

\resizebox{1\linewidth}{!}{  
        \setlength{\tabcolsep}{2.5mm}  
        
        \begin{tabular}{lccccccccc}
        \toprule
        Methods & PSNR $\uparrow$ & LPIPS $\downarrow$ & FID $\downarrow$ & LMD $\downarrow$ & Time   \\
        Ground Truth  & N/A            & 0               & 0              & 0                      & -          \\ \midrule

        w/o T w/o F & 35.35 &  0.0362 & 10.287 & 2.687 & 1.5h \\
        w/o T w F & 35.17 &  0.0173 &  \textbf{9.22} & 2.661 & 2.5h  \\
        w/o S w/o F & \textbf{35.54} &  0.0381 & 10.949 & 2.663 & 1.5h  \\
        w/o S w F    & 35.21    & 0.0172          &  9.550          & 2.662    & 2.5h      \\ 
        
        w T w S w F  & 35.39  & \textbf{0.0169} & 9.472 & \textbf{2.601} & 2.5h  \\
        \bottomrule 
        
        \end{tabular}
    }
    \setlength{\abovecaptionskip}{0cm}
    \vspace{-0.2em} 
    
    \caption{\small \textbf{The quantitative results of  ablation Study of DT-NeRF (Our)}. In obama dataset \cite{guo2021ad} under the resolution of $450\times450$. T: transformer, F: finetune, S: space fusion}
    \label{tab:setting1}
\end{flushleft}  
\end{table}

\vspace{-2.5em} 
\section{Conclusions}
\vspace{-0.4em} 
\label{sec:pagestyle}
In our approach, we propose the use of two decomposed triplane-hash to separately represent the regions of the mouth and the face. Audio features serve as a residual term and are aligned in a coordinate space and the coordinate feature space through an Audio-Mouth-Face Transformer. The aim is to enhance the representational power of the speaking region of the face by optimizing the mouth area. Finally, we leverage the physical representation characteristics of NeRF to augment the representational capability of the facial area through additive volumetric rendering in a unified 3D space. Experimental results validate the superiority of our method.


\vfill\pagebreak


{\small
\bibliographystyle{ieee_fullname}
\bibliography{bib}
}

\end{document}